# Information Access in a Multilingual World

## Proceedings of the SIGIR 2009 Workshop
## July 23, 2009 Boston, Massachusetts USA

**Organized by:**

**Fredric Gey, University of California, Berkeley USA**
**Noriko Kando, National Institute of Informatics, Tokyo JAPAN**
**Jussi Karlgren, Swedish Institute of Computer Science, Stockholm SWEDEN**

**In cooperation with the InfoPlosion Project of Japan:**
**http://www.infoplosion.nii.ac.jp/info-plosion/ctr.php/m/IndexEng/a/Index/**

## Program Committee:

# Introduction and Overview

This workshop is the third workshop on the topic of multilingual information access held during SIGIR conferences this decade. The first was at SIGIR 2002 on the topic of "Cross Language Information Retrieval: A Research Roadmap". The second was at SIGIR 2006 on the topic of "New Directions in Multilingual Information Access". Over the past decade the field has matured and real world applications have appeared. Thus our goal in this 2009 workshop was to collate experiences and plans for the real-world application of multilingual technology to information access. Our aim is to identify the remaining barriers to practical multilingual information access, both technological and from the point of view of user interaction. We were fortunate to obtain as invited keynote speaker Dr Ralf Steinberger of the Joint Research Centre of the European Commission, presenting the Joint Research Centre's multilingual media monitoring and analysis applications, including NewsExplorer. Dr. Steinberger has written an overview paper about their family of applications, which is the first paper in these proceedings.

In our call for papers we specified two types of papers, research papers and position papers. Of the 15 papers initially submitted, two were withdrawn and two were rejected. We accepted 3 research papers and 8 position papers, covering topics from evaluation (of image indexing and of cross-language information retrieval in general), Wikipedia and trust, news site characterization, multilinguality in digital libraries, multilingual user interface design, access to less commonly taught languages (e.g. Indian subcontinent languages), implementation and application to health care. We believe these papers represent a cross-section of the work remaining to be done in moving toward full information access in a multilingual world.

We would like to thank our program committee who read, reviewed and wrote reviews under major time constraints (in some cases 2 days).

Fredric Gey, Noriko Kando and Jussi Karlgren

# TABLE OF CONTENTS

# An introduction to the Europe Media Monitor family of applications

Ralf Steinberger, Bruno Pouliquen & Erik van der Goot
European Commission
Joint Research Centre (JRC)
21027 Ispra (VA), Italy
Tel: (+39) 0332 78 5648
Ralf.Steinberger@jrc.ec.europa.eu

## ABSTRACT
Most large organizations have dedicated departments that monitor the media to keep up-to-date with relevant developments and to keep an eye on how they are represented in the news. Part of this media monitoring work can be automated. In the European Union with its 23 official languages, it is particularly important to cover media reports in many languages in order to capture the complementary news content published in the different countries. It is also important to be able to access the news content across languages and to merge the extracted information. We present here the four publicly accessible systems of the *Europe Media Monitor* (EMM) family of applications, which cover between 19 and 50 languages (see http://press.jrc.it/overview.html). We give an overview of their functionality and discuss some of the implications of the fact that they cover quite so many languages. We discuss design issues necessary to be able to achieve this high multilinguality, as well as the benefits of this multilinguality.

## 1. INTRODUCTION
Most large organizations have dedicated departments that monitor the media to keep up-to-date with relevant developments and to keep an eye on how they are represented in the news. Specialist organizations such as those that monitor threats to Public Health, monitor the multilingual media continuously for early warning and information gathering purposes. In Europe and large parts of the world, these organizations look at the news in various languages because the content found across different languages is complementary. In the context of the European Union, which has 23 official languages, multilinguality is a practical and a diplomatic necessity, and so is the related ability to access information across languages.

Current approaches to link and access related textual content across languages use Machine Translation, bilingual dictionaries, or bilingual vector space representations. These approaches are limited to covering a small number of languages. We present an alternative way of tackling this challenge. It consists of aiming at a language-neutral representation of the multilingual document contents.

The European Commission's *Joint Research Centre* (JRC) has developed a number of multilingual news monitoring and analysis applications, four of which are publicly accessible to the wider public. All four systems can be accessed through the common entrance portal http://press.jrc.it/overview.html.

Users of the systems are the EU Institutions, national organizations in the EU Member States (and in some non-EU countries), international organizations, as well as the public. The freely accessible websites receive between one and two million hits per day from thirty to fifty thousand distinct users.

The next section provides information on the news data that is being analyzed every day. Section 3 gives an overview of the four publicly accessible media monitoring systems and their specific functionalities. Section 4 aims to explain the specific approach which enabled the EMM family of applications to cover a high number of languages and to fuse information extracted from texts written in different languages. It also describes some of the multilingual and cross-lingual functionalities that we could only develop because we adopted this approach. The last section points to future work. Related work for each of the functionalities described will be discussed in the related sections.

## 2. The *Europe Media Monitor* news data
The *Europe Media Monitor* (EMM, see [1]) is the basic engine that gathers a daily average of 80 to 100,000 news articles in approximately 50 languages (status June 2009), from about 2,200 hand-selected web news sources, from a couple of hundred specialist and government websites, as well as from about twenty commercial news providers. EMM visits the news web sites up to every five minutes to search for the latest articles. When news sites offer RSS feeds, EMM makes use of these, otherwise it extracts the news text from the often complex HTML pages. All news items are converted to Unicode. They are processed in a pipeline structure, where each module adds additional information. Whenever files are written, the system uses UTF-8-encoded RSS format.

## 3. The *Europe Media Monitor* applications

The EMM news gathering engine feeds its articles into the four fully-automatic public news analysis systems, and to their non-public sister applications. The major concern of *NewsBrief* and *MedISys* (see [18]) is breaking news and short-term trend detection, early alerting and up-to-date category-specific news display (Sub-section 3.1). *NewsExplorer* (3.2) focuses on daily overviews, long-term trends, in-depth analysis and extraction of information about people and organizations. *EMM-Labs* (3.3) is a collection of more recent developments and includes various tools to visualize the extracted news data.

For NewsBrief and MedISys, there are different access levels, distinguishing the entirely public web sites from an EC-internal website. The public websites do not contain commercial sources and may have slightly reduced functionality.

### 3.1 Live news monitoring and breaking news detection in NewsBrief and MedISys

All EMM news items get fed into NewsBrief and into the *Medical Information System* MedISys as soon as they come in. While NewsBrief is a wide-coverage monitoring tool covering the interests of all users, MedISys sieves out those news reports that talk about potential Public Health threats, including those of chemical, biological, radiological and nuclear origin (CBRN). Every ten minutes and in each of the languages, both applications cluster the latest news items (four hour window or more, depending on the number of recent articles) and present the largest cluster as the current top-ranking media theme (*Top Stories*). The title of the cluster's medoid (the article closest to the cluster centroid) is selected as the most representative title and thus as the title for the cluster. All current clusters are compared to all the clusters produced in the previous round. If at least 10% of the articles overlap between a new cluster and any of the previous ones, the clusters get linked and those articles that have fallen out of the current 4-hour-window get attached to the current cluster. The public web pages are updated every the minutes so that users always see the latest news of the fastest news providers.

Larger new clusters (without overlap to previous clusters) and clusters of a rapidly rising size get automatically classified as *breaking news* so that subscribed users will be notified by email. The statistical breaking news detection algorithm makes use of information on the number of articles and of the number of different news sources, comparing the news of the last 30 minutes with longer periods of time.

Each article is geo-tagged, i.e. potential place names are identified and ambiguities are resolved (for a description of such ambiguities, see Section 0). An algorithm that considers the place hierarchy (city is part of a region which is part of a country) and the frequency of mentions determines the major location in a cluster. This is used to visualize the location of the current news items on a geographical map (see Figure 1).

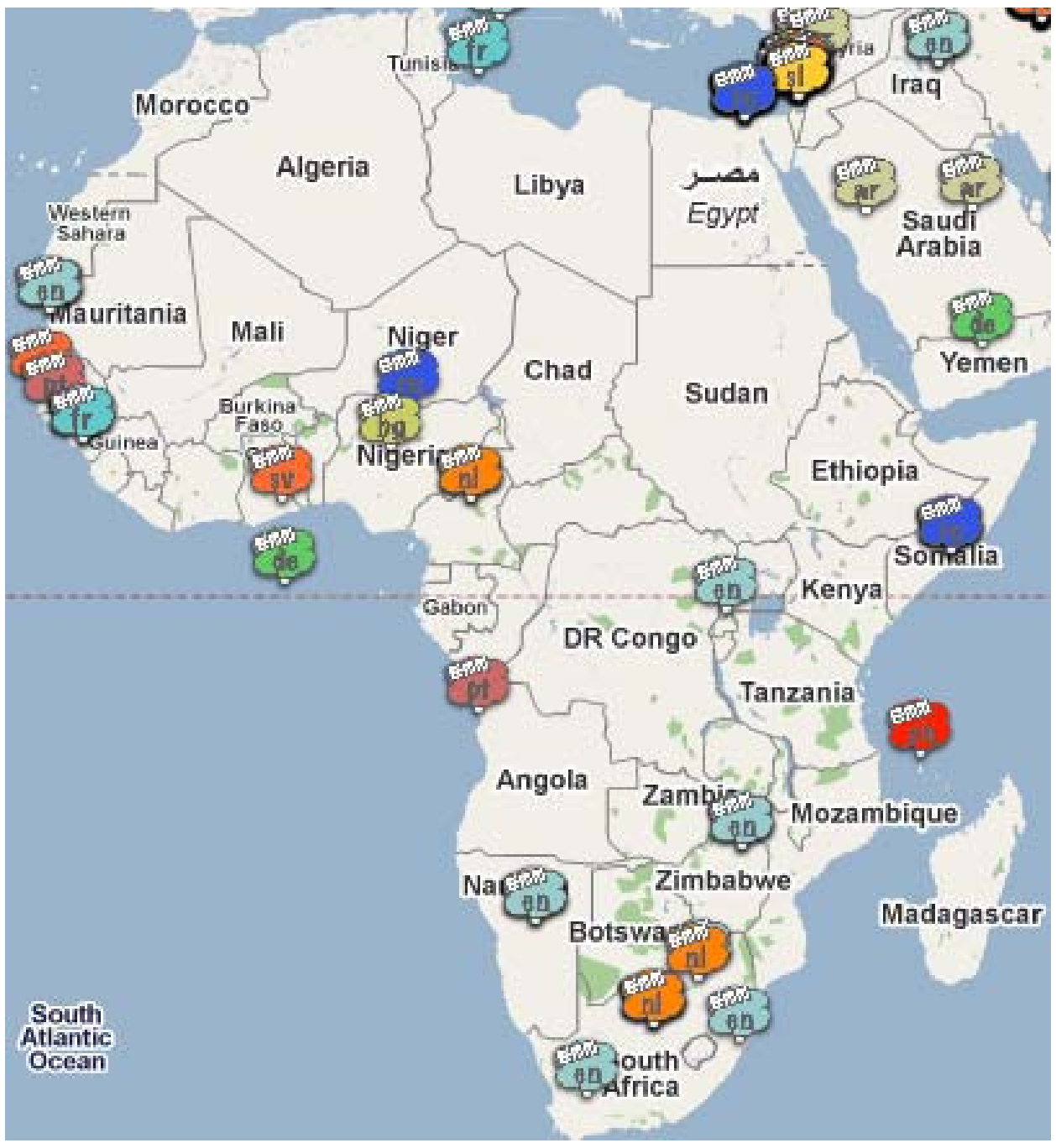

**Figure 1. Map showing the major locations mentioned in news clusters. Different colors indicate different languages.**

All news items are additionally categorized into hundreds of categories. Categories include geographic regions such as each country of the world, organizations, themes such as *natural disasters* or *security*, and more specific classes such as *earthquake*, *terrorism* or *tuberculosis*. Articles fall into a given category if they satisfy the category definition, which consists of Boolean operators with optional vicinity operators and wild cards. Alternatively, cumulative positive or negative weights and a threshold can be used. Uppercase letters in the category definition only match uppercase words, while lowercase words in the definition match both uppercase and lowercase words. Many categories are defined with input from the institutional users themselves.

The system keeps statistics on the 14-day average number of articles falling into any given country-category combination (e.g. *Poland-tuberculosis*). If the number of articles for this combination found in the last 24-hours (normalized by weekday fluctuations) is significantly higher than this average, a country-category-specific alert is triggered and users are notified using alert ranking graphs (see Figure 2).

As the categories are defined in many different languages (depending on the user interests, some are defined in all languages, others in only a few), the statistics are fed by all languages and are thus not language-dependent. This means that the sensitive alerting tool will detect a sudden rise in *any* of the languages, meaning that users may see an alert even before the event is reported in their own language. For humanitarian and public health institutions whose main concern is early warning and rapid reaction, this is a highly appreciated functionality.

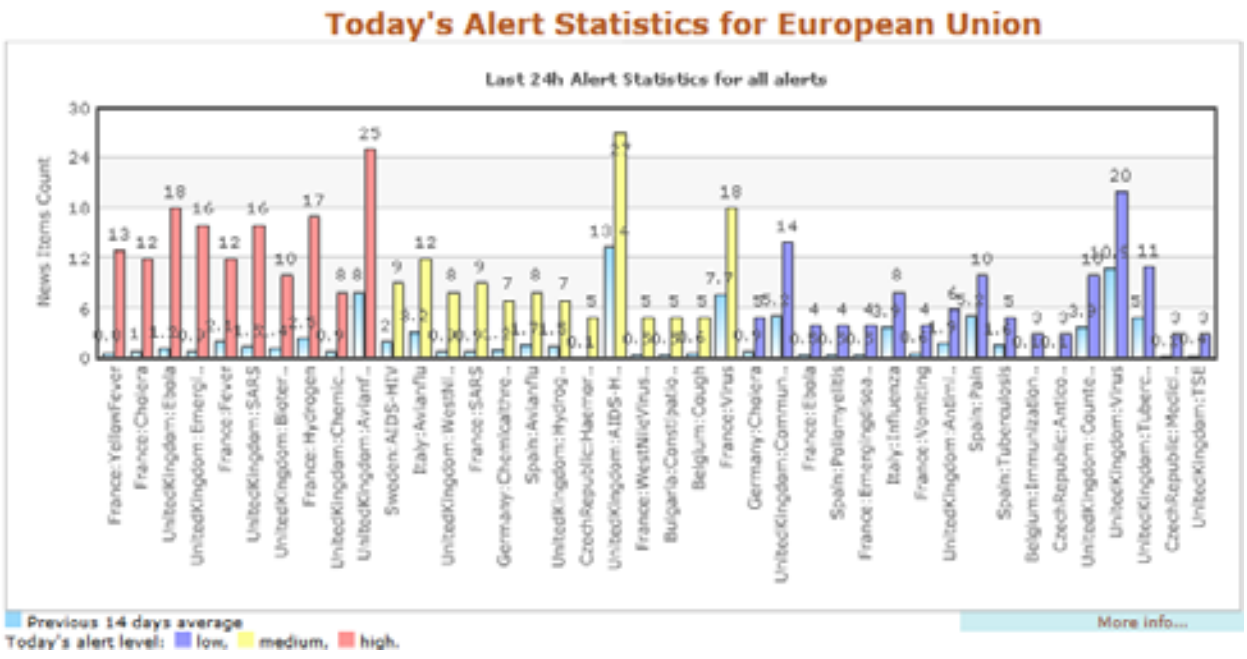

**Figure 2. Graph showing the highest alert levels in the latest news, from 'red' (left) to 'yellow' and 'blue' (right). Each bar stands for a different country-category combination.**

## 3.2 Trend monitoring and information extraction in *NewsExplorer*

NewsExplorer takes as input the non-commercial EMM articles in 19 languages collected within a calendar day, clusters them separately for each of its languages and displays the major news per day, ordered by cluster size. Languages covered include 14 EU languages, and additionally Arabic, Farsi, Russian, Turkish and Norwegian. NewsExplorer applies a number of text mining tools to detect names of persons, organizations and locations and displays those names found in the course of each day. For each cluster, a dedicated web page gets created, where the same type of information is displayed specifically for this cluster. It furthermore detects reported speech quotations by people (*X said "…"*) and about people (*Y said "…X…"*), as well as titles of persons used in the media (e.g. *former foreign minister, playboy, 58-year-old*). Many persons are referred to using variant spellings, not only across writing systems (e.g. Greek *Γκέρχαρντ Σρέντερ*, Russian *Герхард Шредер*, and German *Gerhard Schröder* all refer to the former German Chancellor) and across languages (e.g. English *Vladimir Ustinov* and German *Wladimir Ustinow*), but also within the same language. For instance, the following variants were found within English language news:

| Names |
|---|
| Barack Obama (Eu,yo) |
| Barak Obama (az,wo) |
| Барак Обама (ba,uk) |
| باراك أوباما (ar) |
| باراك اوباما (ar,fa) |
| باراک اوباما (fa) |
| Barack Hussein Obama (da,tr) |
| Barrack Obama (da,tr) |
| Obama Barack (da,sv) |
| Barack Hussein Obama II (an,tt) |
| Barack Hüseyin Obama (tr) |
| Barac Obama (en,tr) |
| Barack Hussain Obama (en) |
| Barrack Hussein Obama (en,fr) |
| Barrak Obama (en,tr) |
| Барак Хуссейн Обама (ru) |
| Baraque Obama (pt) |
| バラク・ オバマ (ja) |

| Names |
|---|
| บารัค โอบามา (th) |
| Բարաք Օբամա (hy) |
| ބަރަކް އޮބާމާ (dv) |
| Baracus Obama (la) |
| Ɔbenem (ak) |
| באראק אבאמא (yi) |
| 贝拉克·奥巴马 (zh) |
| ברק אובאמה (he) |
| ބަރަކް އޮބާމާ (dv) |
| Baraks Obama (lv) |
| Барак Хусейн Обама (ru) |
| Обама Барак (uk) |
| 버락 오바마 (ko) |
| ಬರಾಕ್ ಒಬಾಮ (kn) |
| பராக் ஒபாமா (ta) |
| बराक ओबामा (hi,mr) |
| بارک اوباما (ur) |
| Barack Hüssein Obama (tr) |

**Figure 3. Some of the name variants found for Barack Obama.**

*Schröder*, *Schroeder*, *Schröeder*. NewsExplorer aims to detect these variants as belonging to the same person. It additionally retrieves more name variants (e.g. Chinese, Japanese and Thai) from Wikipedia, resulting in up to 170 name variants for the same person. Figure 3 shows some name variants for the current US President. Users searching the system for any of these will also find all articles using any of the alternative spellings. As of June 2009, the name database contains 900,000 names plus about 170,000 variants. Names from the name database are exported daily into a finite-state automaton so that these known names will also be found in EMM's live news monitoring systems described in Section 3.1.

For each person or organization name (referred to as *entity*) detected in at least two different news articles in the same cluster, NewsExplorer displays all the extracted meta-information on individual news pages, one per entity. NewsExplorer also computes which entities get frequently mentioned together (displayed as *related names*) and – more interestingly – which entities get mentioned frequently with the first person *but do not have much media attention for themselves*, using a TF.IDF-like formula (see [15]). This latter group of persons, referred to as *associated names*, typically consists of close contacts such as relatives, personal spokespersons or secretaries, etc. An interactive visualization tool allows users to display entities, associated names, and names jointly associated to various entities (See Figure 4).

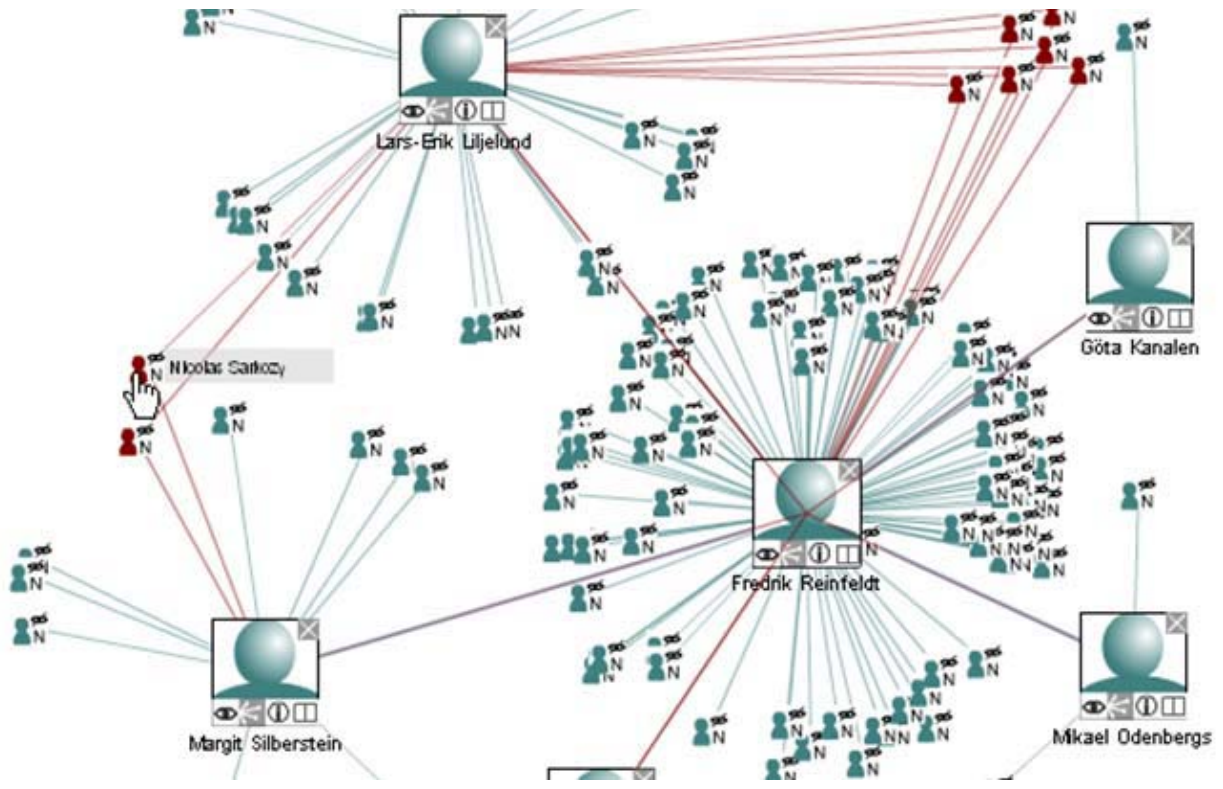

**Figure 4. Visualizing relations between *associated people*.**

For all the daily clusters, NewsExplorer establishes and displays links to related clusters found in the previous days. Many events or themes capture the media's attention for a longer period, such as the Israel-Palestine or the Iran-nuclear conflicts. NewsExplorer links all these historically related news clusters into *stories*. For each story, an interactive graph visually presents media reporting activity over time (Figure 5) and displays story-specific meta-information, such as names, countries and keywords extracted from all the – sometimes thousands – related news articles.

Together with the name variant mapping tool (see also 4.3.3), the probably most interesting feature of NewsExplorer is the cross-lingual cluster linking functionality: for each cluster, users can – with a simple click on the mouse – jump straight to the equivalent news reports in the other languages. NewsExplorer establishes links between clusters in different languages if the multilingual clusters have a minimum cross-lingual similarity based on subject domains, person and organization names, locations and cognates (words that are the same across languages). This application is described in Section 4.3.1.

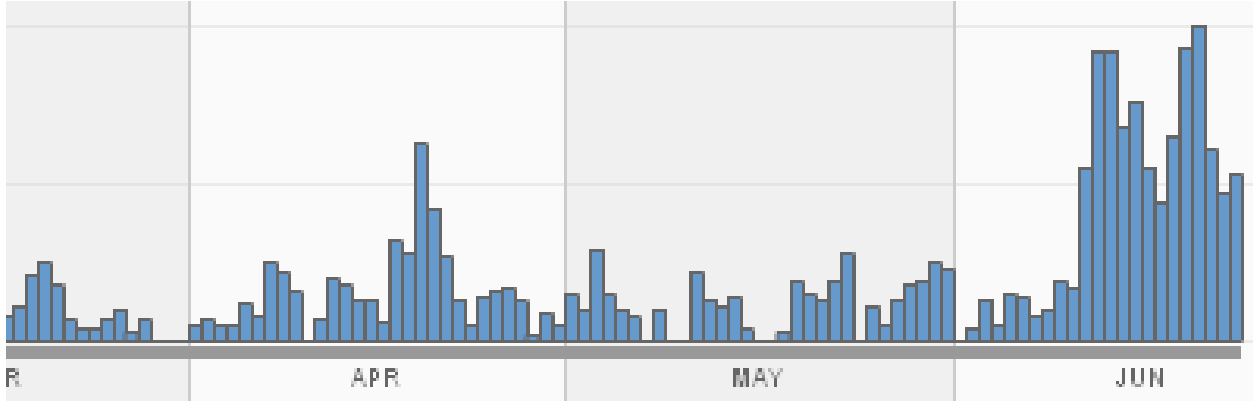

**Figure 5. Interactive graph showing a 'story' timeline on the 2008 Iran elections. Numbers of daily articles increased drastically when the post-election clashes broke out in June.**

## 3.3 Further visualization and analysis tools in *EMM-Labs*

EMM-Labs is a loose collection of further media-based text mining and interactive visualization tools. It includes generic or topic-related geographical news maps (e.g. on *swine flu*), various types of statistics on EMM categories over time and across languages, and social network browsers showing generic co-occurrence frequency relations between persons or specific extracted relations such as *support*, *criticism* and *family* relations.

The news cluster navigation tool allows an interesting alternative view of the news: it graphically shows an interactive network of the latest news clusters, the most active categories at the moment, and the countries most mentioned in the news right now, with stronger and weaker links between the nodes belonging to any of the three groups (see Figure 6). The interactive interface of the tool lets users decide on various thresholds (numbers of clusters displayed, percentage of articles of a cluster that need to be classified according to country or category for the link to be displayed, etc.).

The event extraction tool detects violent events, accidents, natural and humanitarian disasters and more in the live news clusters (fed by NewsBrief) in the languages English, Russian, French, Italian and Spanish, as well as in Arabic (after an analysis of the English text machine-translated from Arabic). The tool detects and displays information on the event type, on the location of the event, as well as on number and status of the victims (see Figure 7 and Figure 8). The displayed facts are a best-guess combination of the information found in all the articles of the news cluster. For details on this application, see [2].

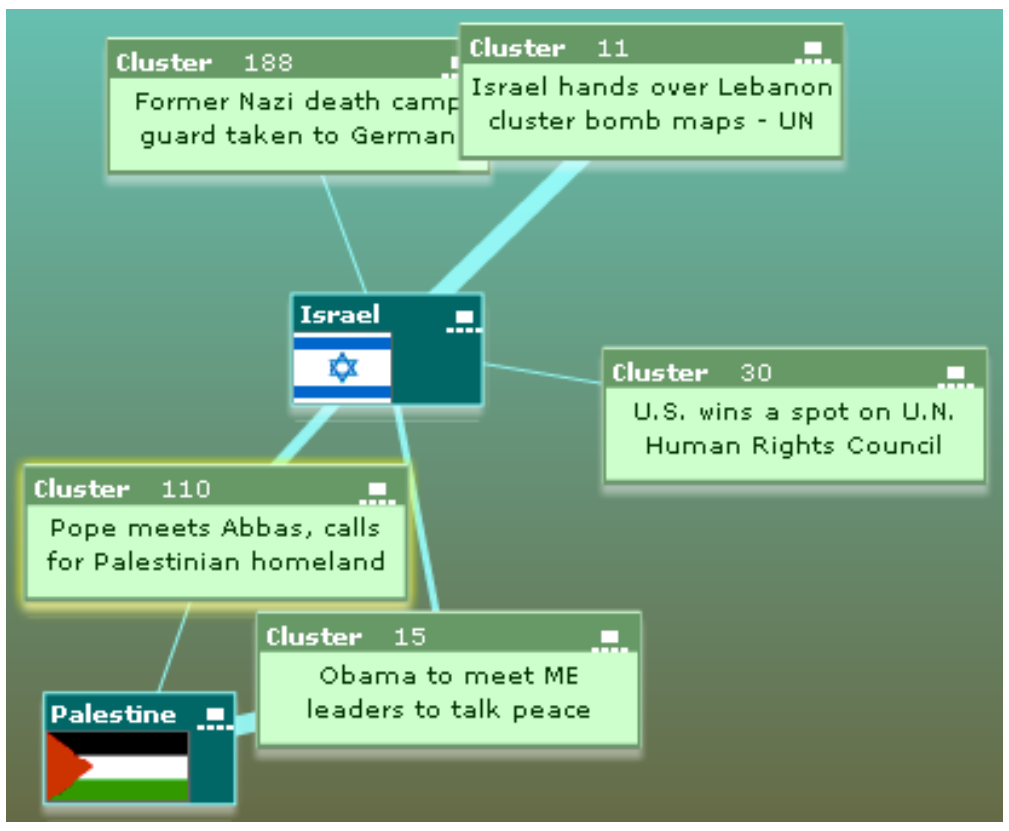

**Figure 6. Cluster browser visually presenting current news clusters and related country categories.**

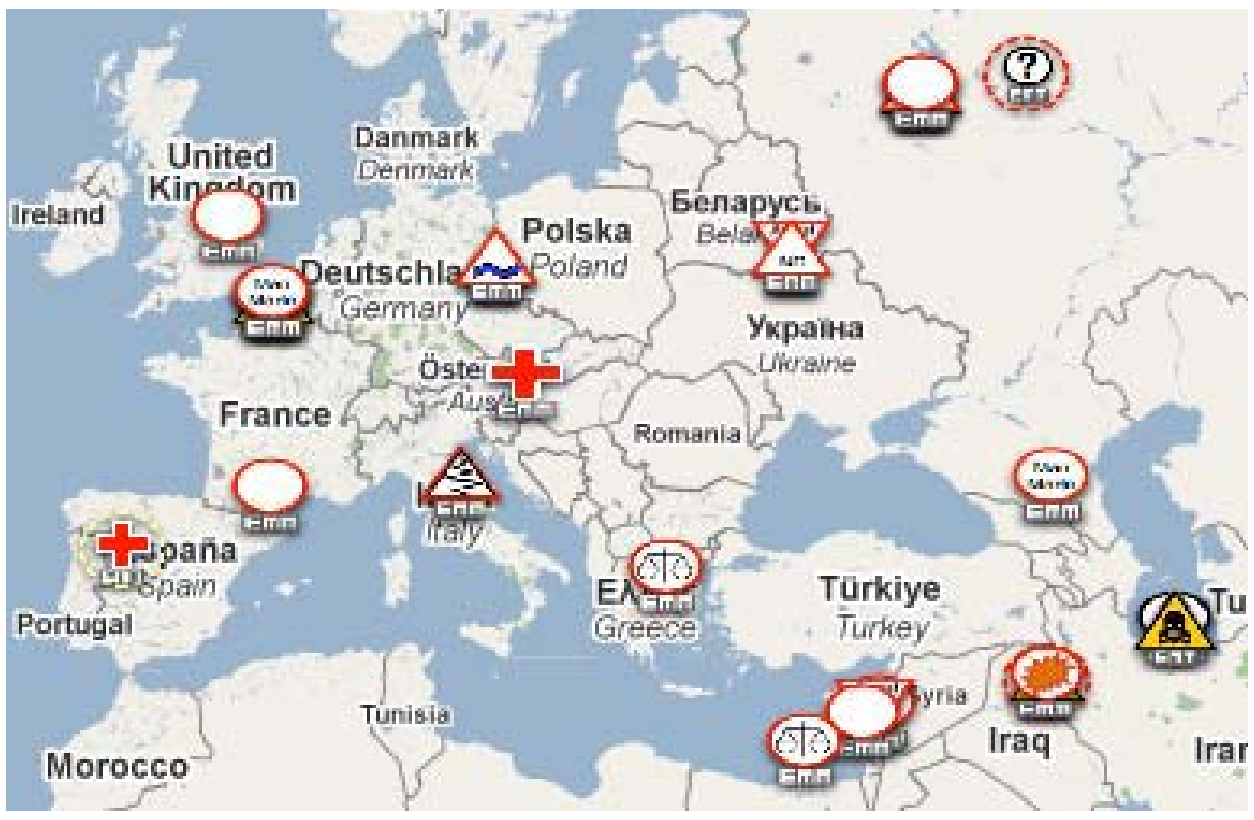

**Figure 7. Map showing events involving victims, extracted from the news in 6 languages.**

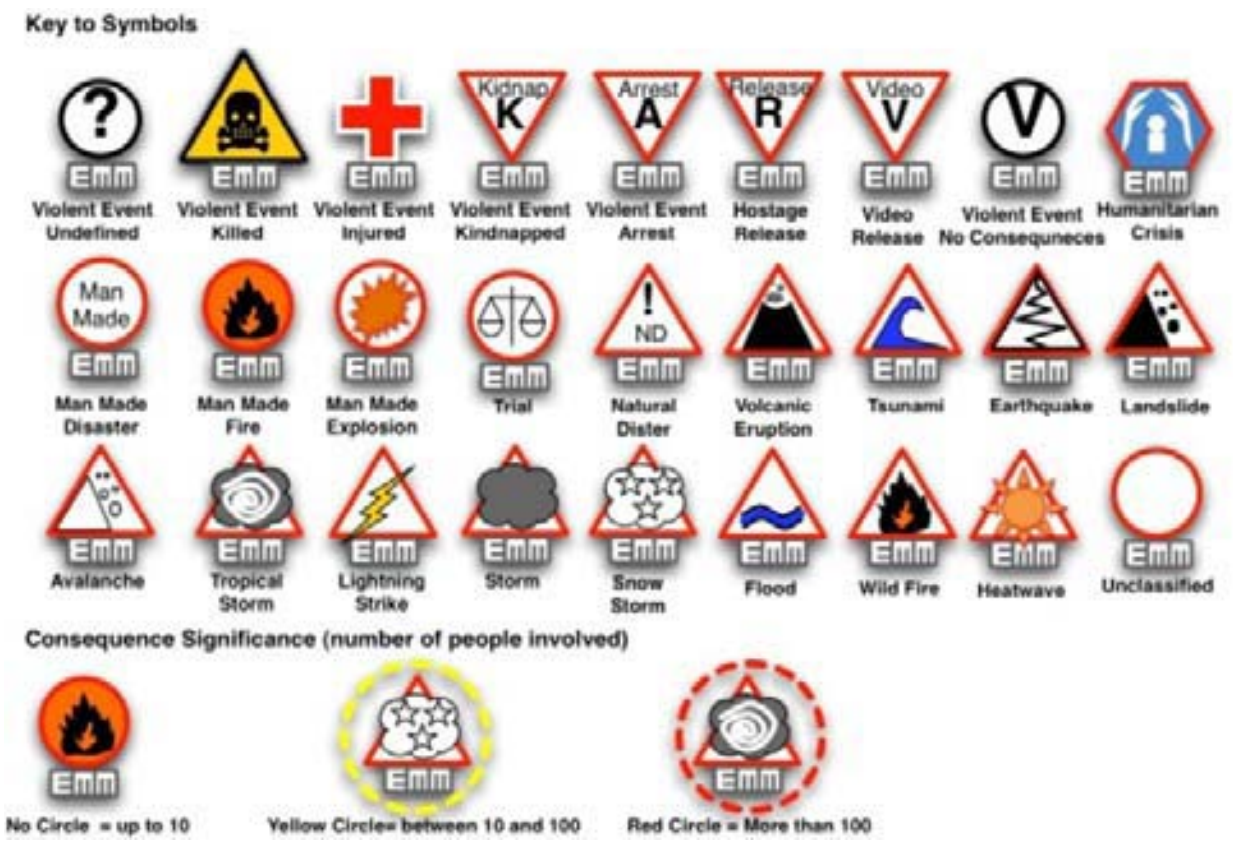

**Figure 8. Event types recognized by the system.**

# 4. Approach to multilingual and cross-lingual information processing

Since their conception, the prerequisite for EMM applications was that they should be highly multilingual, aiming at all EU languages and possibly more. This challenging objective was to be achieved with a relatively small-sized developer team, so that it was clear that any components and processes needed to be simple and, ideally, language-independent. While – to some extent – the simplicity was a hindrance as it did not allow to consider many language-specific features, the achieved multilinguality brought many advantages not usually achieved by systems covering one or only a few languages. In this section, we want to highlight some of the methods used and describe some of the benefits of multilinguality in EMM.

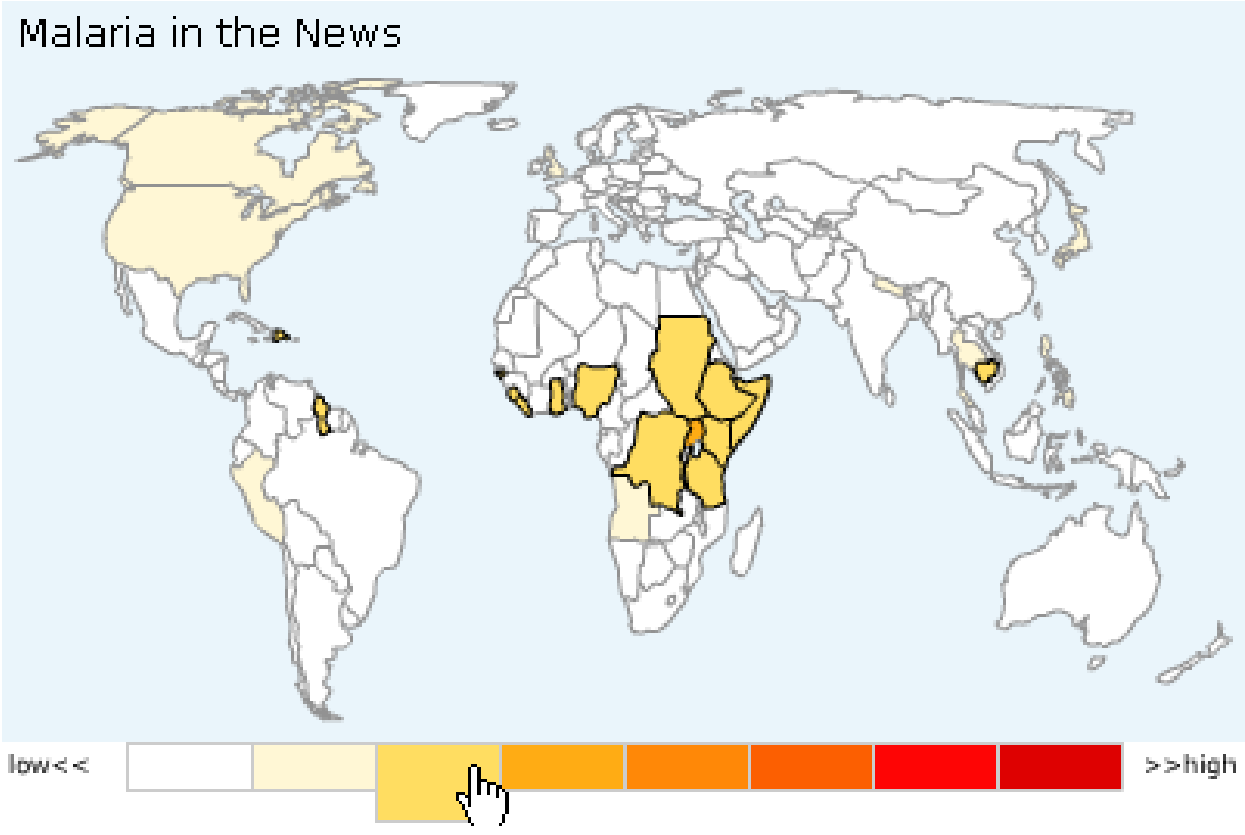

**Figure 9. World map showing countries mentioned in the context of Malaria.**

## 4.1 Country and category classification across languages

Category definitions in *NewsBrief* and in *MedISys* are multilingual. News items from different languages thus get categorized into exactly the same classes. While we use the English name as the class label, the classes themselves are thus being filled with news articles from many different languages. This has the advantage that trend graphs showing an increase in articles per category may alert users about increased activity in a specific field *even before news articles in their own language are published* (see Figure 2 and the explanations in Section 3.1). The categorization of texts according to the countries mentioned is treated in exactly the same way, i.e. there is one EMM class for each country of the world. This allows to display multilingual country-category information (the intersection of any country category with any subject domain category) on maps (see Figure 9). Note that the geo-tagging application described in Section 4.2 is performed independently of this country categorization.

The method to categorize the articles, described in Section 3.1, is rather simple and user-friendly, and it lends itself to dealing with many languages. EMM uses only one definition per category for all languages combined, meaning that English, French, Russian, etc. definition words are all mixed inside the same definition and are thus all searched in all languages. This has the obvious disadvantage that false friends (same words meaning different things across language) may sometimes lead to false positives. For instance, users searching for the disease symptom '*pain*' may erroneously receive French articles about bread as the French word '*pain*' means *bread*. However, these multilingual definition lists also have advantages: (a) They are very easy to maintain – which is not a negligible criterion when dealing with hundreds of categories in about 50 languages; (b) they automatically benefit from cognate words that are the same across many languages (e.g. *tsunami*), or that are almost the same, so that one simple search word with wildcards will often allow to match the words in various languages (e.g. *tuber_ul%* will match English *tuberculosis*, German *Tuberkulose*, French *tuberculose* and *tuberculeux*, etc.). The obvious alternative to producing hand-crafted category definitions would theoretically be the usage of machine learning methods for categorization (e.g. using Support Vector Machines or Naïve Bayes), but most EMM users would not be willing to provide training examples in all the languages. For many categories, such as diseases, organizations or subjects (such as *genetically modified organisms*, GMO) it is much easier to gather multilingual search terms from Wikipedia or to ask colleagues for help who speak the languages of interest. Users regularly request the monitoring of new news categories, and due to the simplicity of the approach, first versions of the category definitions can often be produced within a day.

## 4.2 Multilingual information extraction using mostly language-independent rules

Developing text mining components such as tools to extract structured information from unstructured free-text typically requires writing (grammar) rules and using dictionaries. Developing such tools for many languages is rather labor-intensive. Developing time can be saved by re-using such linguistic resources of one language to develop the resources for the other languages (see, e.g. [4], [10], [9]). However, even adapting existing resources to new languages is still rather time-consuming. In NewsExplorer, which carries out Information Extraction (IE) for 19 languages, we thus aimed at using the same, language-independent rules for all languages, and to store any language-dependent information in language-specific parameter files. From the conception phase of any text analysis tool development in NewsExplorer, the multilinguality requirement is an unmovable premise. For details, and for examples how these principles were realized in eight different text analysis applications, see [20].

A simplified example of such a language-independent rule is the quotation recognition pattern below, which extracts reported speech and the person issuing the quotation, by making use of the slots *name*, *reporting-verb* and *quote-mark*. The set of possible quote marks (such as <<, >>, “, ”, etc.) is mostly the same across languages. The English language-specific parameter file includes lists of reporting verbs such as *said*, *reported*, *added*, etc. and their morphological variants (e.g. *says*, *was saying*). Person names are identified and marked up using a separate tool, which is based on the same principles. Further slots (not shown here) may include *modifier* and *auxiliary-verb* for such a rule. Elements within square brackets are optional.

> *name* [, *up to 60 chars* ,] *reporting-verb* [:|that] *quote-mark*
> QUOTE *quote-mark*
> e.g. *John Smith, supporting AFG, said: "They are the best!".*

Another example for language-independent rules is the geo-tagging application, which includes geo-parsing (identification of potential geographic references in text) and disambiguation. Geo-tagging requires a gazetteer (containing lists of locations and their geo-co-ordinates), which is thus an unavoidable linguistic resource. Disambiguation is necessary due to various types of homography, i.e. between locations and persons (*Paris* in France vs. *Paris Hilton*), between locations and generic words (e.g. *Split* is a city in Croatia) and between locations which share the same name (e.g. there are 15 locations world-wide called *Paris* and there are 32 places called *Washington*). The geo-disambiguation rules in NewsExplorer are language-independent, as they make use of features such as the information whether an entity has previously been tagged as being a person or organization name, gazetteer-provided information on the size class of the locations (capital, major city, city, town, …), information on the country of origin of the news source, information on other potential geographic references in the same text, as well as kilometric distance between the ambiguous place and other, non-ambiguous places in the same text. See [13] for details. Apart from the gazetteer, the only language-specific information is a geo-stop word list, which can be created within a few hours of work. Such a geo-stop word list is used to exclude false positives for locations like *And* and *By* (locations in Iran and in Sweden, respectively), which are homographic with high frequency words of a language. By keeping language-specific information out of the disambiguation rules, any new language can be plugged into the system if a gazetteer for this language is available.

## 4.3 Cross-lingual information fusion across over one hundred language pairs

Examples for cross-lingual information access applications are cross-lingual information retrieval (CLIR), cross-lingual information access (CLIA, e.g. cross-lingual glossing, or the linking of related documents across languages), cross-lingual name variant matching (identifying that *Ali Chamenei*, *Alí Jamenei* and *Али Хаменеи* are the same name), cross-lingual plagiarism detection, multilingual summarization and machine translation (MT). Known approaches to these applications make use of MT (e.g. [8]) or of different kinds of bilingual dictionaries or thesauri (manually produced or machine-generated, e.g. [24]). An alternative approach consists of producing bilingual word associations by feeding pieces of parallel text to systems applying Lexical Semantic Analysis (LSA, [6]) or Kernel Canonical Correlation Analysis (KCCA, [23]). All of these approaches have in common that they are basically bilingual ([3] being a notable exception), meaning that multilingual applications are actually multi-bilingual. In a highly multilingual environment with N languages, there are vast numbers of language pairs, namely: `N*(N-1)/2`. In NewsExplorer, which covers 19 languages, there are thus 171 language pair combinations, and 342 language pair directions. A multi-bilingual approach clearly is impractical.

### *4.3.1 Components used in NewsExplorer*

For NewsExplorer's components that link related documents across languages or that fuse information found in different languages, the only possible solution was thus to work with an interlingua type of approach, i.e. representing contents in a language-neutral way and mapping contents in each of the languages (news articles, news clusters, or extracted pieces of information) to this language-independent representation. The following representations are used in NewsExplorer (for details, see [14]): (a) A vector of multilingual subject domain representations using the Eurovoc thesaurus (see [16] and Section 4.3.2); (b) a frequency list of locations per document, represented by location identifiers and their latitude-longitude information (see [13] and Section 4.3.3); (c) a frequency list of person and organization name identifiers (also described in Section 4.3.3); (d) monolingual weighted lists of keywords. Identifiers (b) and (c) each represent a whole range of multilingual variants. Sections 4.3.2 and 4.3.3 give details on these four ingredients for cross-lingual cluster linking.

Further possible alternatives for a language-neutral representation (not currently exploited), are normalized expressions to express dates or measurement units (speed, acceleration, time, etc.), subject-specific multilingual nomenclatures such as the *Medical Subject Headings* MeSH or the *Customs Tariff Code* TARIC, and more (see [19] for further ideas and more detail).

### *4.3.2 Multilingual subject domain representation*

The idea of ingredient (a) in 4.3.1 was thus to represent the contents of documents by a ranked list of subject domains (also referred to as *topic maps*), using a multilingual classification of subject domains that are the same across languages. The only resource we are aware of that is available to be used for such a purpose is Eurovoc[1]. Eurovoc is a wide-coverage thesaurus with over 6,000 classes that was developed for human subject domain classification in the European Commission and in parliaments in the European Union and beyond. As Eurovoc has been used for many years, tens of thousands of documents exist that have been labeled manually by librarians. The documents in 22 languages which are part of the freely available multilingual parallel corpus JRC-Acquis ([21]) are accompanied by their respective Eurovoc codes. The classifier that assigns Eurovoc codes to news clusters in NewsExplorer was developed by applying machine learning techniques, using these manually classified documents. The major challenges of this task were that this is a multi-label task (each manually classified document belongs on average to 5.6 classes) involving a large number of categories (more than 3,000 of the 6,000 Eurovoc classes are actively used) with a highly imbalanced category distribution and heterogeneous text types. The adopted approach, described in [16], produces a ranked and weighted list of (numerical) Eurovoc subject domain codes. In NewsExplorer, the cosine similarity is applied to compare this language-independent subject domain representation for the daily news clusters in one language with those in other languages. This is the major ingredient for the cross-lingual cluster similarity measure in NewsExplorer.

### *4.3.3 Language-neutral entity vectors*

The second most important ingredient for cross-lingual cluster linking in NewsExplorer ((b) in 4.3.1) is the similarity based on geo-references. We described in Section 4.2 the principles behind our geo-tagging application, which recognizes potential location names in free text and resolves any ambiguities using language-independent rules. This tool produces, for each document in each of the languages, a frequency list of locations including the country they belong to. At cluster level, NewsExplorer aggregates this information by counting the references to each country (e.g. one mention of the Turkish city of *Izmir* will add one count to the country count of *Turkey*, which is itself represented by the language-neutral country ISO code) and produces a frequency count of (direct or indirect) country mentions. The cosine similarity is applied to the country vector representation of each pair of clusters in the different languages to calculate the value of the second ingredient.

The third ingredient vector ((c) in 4.3.1) consists of a frequency list of normalized person or organization names, where the multilingual name variants for the same entity are all represented by one language-neutral numerical code (see Figure 3). Named-entity recognition rules are applied to all the news in all the NewsExplorer languages. The rules were developed according to the principles laid out in Section 3.2. These rules identify name mentions in each of the languages. If the name variants found are already known, the names will be represented by the language-neutral entity identifier. If the name mention found is a previously unknown name, a name variant matching procedure will be launched. This procedure first transliterates non-Latin script names into the Latin character set, using standard hand-crafted transliteration rules. It then normalizes all names by applying about 30 different rules, which try to map language-specific spelling differences onto one canonical form. Examples for such empirically found differences are, for instance, that diacritics of foreign names are often omitted (e.g. outside Poland, the name *Lech Wałęsa* is usually spelt as *Walesa*) and that consonants are frequently doubled or singled (e.g. *Mohammed* vs. *Mohamed* and *Barack* vs. *Barrak*, see Figure 3). Russian names ending in *-ов* (*-ov*) may furthermore be spelled as *-ev*, *-ow*, *-ew*, etc., depending on the target language of the transliteration. After applying these normalization steps, which aim at mapping frequent spelling variations to one canonical spelling, all vowels are removed because vowels often differ, especially for transcriptions of Arabic names. The result of the name normalization is the canonical form of this name (e.g. `mhmd hmdnjd` for *Mahmoud Ahmadinejad*, and `kndlz rc` for *Condoleezza Rice*). The canonical form of each newly found name is then compared to the canonical form of all known names and their variants. If the two canonical forms are the same, the Levenshtein edit distance is applied to two different representations of the name: (1) the found surface string of a name or the transliteration result, and (2) the normalized form with vowels. If the combined similarity between two names is above an empirically established threshold, the name variants get automatically merged and the new name becomes a known name variant of an existing name. In the course of five years of daily analysis, NewsExplorer has altogether identified over one million names and name variants, with as many as 170 different name variants for the same person. For a detailed description of this name variant matching approach, see [12]. Unlike other known methods to map names across scripts ([7], [5]) or to compute string similarity measures ([11]), the presented algorithm is not multi-bilingual, but uses a single representation for all languages. New languages can either be added without any additional steps, or one or more new normalization rules can be added to the set of normalization rules. All normalization rules are applied to all languages. For the purposes of creating the third ingredient for the cross-lingual cluster similarity calculation, each news cluster is represented by a frequency list of all known or new names, with all name variants being represented by the same numerical entity identifier. Each daily news cluster will thus again be compared to all the clusters in the other languages, using the cosine similarity measure.

[1] See http://europa.eu/eurovoc/.

The fourth ingredient for cross-lingual cluster similarity is simply a log-likelihood-weighted monolingual list of words found in each cluster. The cosine similarity is again applied to compare the vector representations in the different languages. While most of the words are of course different across languages, this fourth similarity still adds to the other three similarities as it considers cognates, names, name parts, numbers and acronyms.

The four ingredients are combined with a relative weight of 0.4, 0.3, 0.2 and 0.1. These weights were set intuitively and after comparing various examples manually, but they have not yet been confirmed empirically. See [14] and [20] for more detail.

### *4.3.4 Fusion of information about entities found in documents written in different languages*

The EMM system's information extraction components extract different types of information about persons and organizations, including: (a) name variants, (b) name attributes such as titles (e.g. *president*, *stuntman*, *58-year-old*), (c) news clusters in which the entity was mentioned, (d) stories in which the person was mentioned, (e) quotations by and about the person, (f) co-occurrence information (pure frequency and a weighted value) for entities being mentioned in the same clusters, (g) labeled relations between persons (English only), namely relations of *support*, *criticism*, *contact* and *family relation*. EMM applications produce daily and long-term social networks of different types ([17], [22]), based on weighted co-occurrence, based on labeled relations, and based on who mentions whom in reported speech.

As each entity (independently of the name spelling) is represented by one numerical identifier, the information extracted from any of the input languages can be displayed together. To see examples of information gathered across many languages about the same entity, go to NewsExplorer and click on any of the entity names. An additional benefit of having captured the multilingual name variants is that the generated social networks are fed by news articles in many different languages, so that they are independent of the reporting bias by one country or reporting language. Instead, they represent a rather rounded multilingual and multi-national picture of the relations between persons.

## 5. Summary and Future Work

We have given an overview of the functionality of the EMM family of media monitoring applications (Section 3) and we tried to highlight two issues regarding the fact that EMM covers many different languages (Section 4). First, working on many languages while being a small team automatically forced us to using simple means. We tried to show how such simple means could be used to develop relatively complex and powerful applications. Second, we tried to give some examples of a number of applications benefitting from this multilingual news processing.

Currently ongoing work focuses on (a) ensuring a tighter integration of the various applications and tools, (b) adding opinion mining functionality to the existing information extraction components, (c) producing multi-document summaries of the various news clusters in order to allow notifying users more efficiently of breaking news events, and (d) adding blogs as a new text type to the current news monitoring systems. Like for the existing applications, the challenge is to find methods that can be applied to many different languages.

## 6. ACKNOWLEDGMENTS

The EMM applications have been developed over many years, by many people. We want to thank all contributors for their dedicated work. The following persons should be mentioned explicitly: Flavio Fuart helped to make the EMM output available on robust web pages; Martin Atkinson produced several of the graphical tools, and Jenya Belyaeva contributed with her knowledge of many languages and her thorough quality control.